\documentclass[conference]{IEEEtran}
\IEEEoverridecommandlockouts
\usepackage{cite}
\usepackage{amsmath,amssymb,amsfonts}
\usepackage{algorithmic}
\usepackage{graphicx}
\usepackage{textcomp}
\usepackage{xcolor}
\usepackage{booktabs}
\usepackage{threeparttable}
\usepackage{multirow} 
\usepackage{makecell}
\usepackage{soul,color}
\usepackage{booktabs}
\usepackage{tabularx}
\usepackage{array}
\usepackage{float}

\usepackage{url,hyperref,microtype}
\hypersetup{
    colorlinks=true,
    citecolor=blue,
    linkcolor=blue,
    filecolor=magenta,      
    urlcolor=black
}

\def\BibTeX{{\rm B\kern-.05em{\sc i\kern-.025em b}\kern-.08em
    T\kern-.1667em\lower.7ex\hbox{E}\kern-.125emX}}
\begin{document}

\title{Conference Paper Title*\\
{\footnotesize \textsuperscript{*}Note: Sub-titles are not captured in Xplore and
should not be used}



\title{Multimodal BVP and Respiration for Automated Pain Recognition with Hankel-FCN}


}


\makeatletter
\newcommand{\linebreakand}{%
  \end{@IEEEauthorhalign}
  \hfill\mbox{}\\
  \mbox{}\hfill\begin{@IEEEauthorhalign}
}
\makeatother
\author{
\IEEEauthorblockN{Valentina Becchetti}
\IEEEauthorblockA{\textit{Department of Computer, Control}\\
\textit{\& Management Engineering} \\
\textit{Sapienza University of Rome} \\
Rome, Italy \\
v.becchetti@diag.uniroma1.it}
\and
\IEEEauthorblockN{Muhammad Umar Khan}
\IEEEauthorblockA{\textit{BioSIS (Biosensing \& Intelligent Systems) Lab}\\ 
\textit{Centre for Intelligent Computing \& Systems} \\
\textit{University of Canberra} \\
Canberra, ACT, Australia \\
umar.khan@canberra.edu.au}
\and
\IEEEauthorblockN{Stefanos Gkikas}
\IEEEauthorblockA{\textit{Honda Research Institute Japan} \\
Wako City, Japan \\
stefanos.gkikas@jp.honda-ri.com}
\linebreakand
\IEEEauthorblockN{Alessandro Giuseppi}
\IEEEauthorblockA{\textit{Department of Computer, Control}\\
\textit{\& Management Engineering} \\
\textit{Sapienza University of Rome} \\
Rome, Italy \\
giuseppi@diag.uniroma1.it}
\and
\IEEEauthorblockN{Raul Fernandez Rojas}
\IEEEauthorblockA{\textit{BioSIS (Biosensing \& Intelligent Systems) Lab}\\ 
\textit{Centre for Intelligent Computing \& Systems} \\
\textit{University of Canberra} \\
Canberra, ACT, Australia \\
raul.fernandezrojas@canberra.edu.au}
}

\maketitle

\begin{abstract}
Automated pain recognition is a fundamental challenge for the advancement of clinical care and empathetic human-computer interaction, due to the subjective and dynamic nature of pain. In this study, we propose a deep learning framework for objective pain assessment based on multimodal physiological signals from the AI4Pain dataset. The approach focuses on Blood Volume Pulse (BVP) and Respiration (RESP), capturing complementary aspects of autonomic responses to nociceptive stimulation. To explicitly model their temporal dynamics, the signals are transformed into structured trajectory representations using Hankel matrix embeddings, enabling the learning of dynamical patterns beyond conventional time-series representations. These embeddings are processed by modality-specific Fully Convolutional Networks, and their outputs are integrated through a late fusion strategy for three-class classification (No Pain, Low Pain, High Pain). Experimental results demonstrate that the proposed method achieves an accuracy of 67.09\% and an F1-score of 66.41\%, providing competitive performance compared to existing approaches, while relying on no handcrafted feature engineering. Multimodal fusion significantly outperformed either modality alone, and the Hankel embedding showed a trend towards further improvement. Further validation is required to establish generalisability to broader populations and clinical settings.
\end{abstract}


\section{Introduction}


Automatic pain recognition remains a significant challenge in clinical care and empathetic human–computer interaction due to the subjective and dynamic nature of pain. Conventional pain assessment methods primarily rely on self-report scales or behavioral observations \cite{payen2001assessing,urden2013critical}, which are inherently subjective and unsuitable in situations where patients are unable to communicate effectively, such as during sedation, coma, or severe neurological impairment. Furthermore, pain is not a static phenomenon but a dynamic physiological process involving interactions between peripheral and central nervous system mechanisms \cite{fernandez2023systematic}. These characteristics motivate the development of objective and continuous approaches capable of capturing the complex physiological manifestations of pain.

Physiological signals provide objective measurements of autonomic nervous system activity associated with nociceptive stimulation \cite{afenigus2024evaluating}. Signals such as Blood Volume Pulse (BVP), Respiration (RESP), Electrodermal Activity (EDA), and Peripheral Oxygen Saturation (SpO$_2$) capture complementary physiological responses related to cardiovascular activity, respiratory modulation, sympathetic arousal, and peripheral perfusion \cite{khan2025systematic}. These modalities offer real-time insights into physiological changes that are often imperceptible through behavioral assessment alone, making them particularly suitable for automated pain monitoring systems.

Despite the advantages of physiological sensing, relying on a single modality is often insufficient to capture the full complexity of pain. Pain emerges from the coordinated interaction of multiple physiological systems, and each modality reflects only a partial representation of the underlying nociceptive process \cite{garland2012pain}. For example, cardiovascular responses may reflect autonomic arousal, while respiratory signals capture alterations in breathing dynamics under pain conditions \cite{fernandez2023multimodal}. Consequently, multimodal physiological analysis has gained increasing attention as a means to integrate complementary sources of information and provide a more robust and comprehensive representation of pain-related dynamics \cite{gkikas2023automatic}.

Traditional machine learning approaches for physiological pain assessment typically rely on handcrafted feature extraction \cite{fernandez2019machine}. Although statistical and frequency-based descriptors can capture certain signal characteristics, they are often limited in their ability to represent the complex temporal and non-linear relationships inherent in physiological responses to pain \cite{nargesian2017learning, kyprakis_gkikas_acii_2026}. Deep learning offers a powerful alternative by enabling the automatic extraction of hierarchical features directly from raw or minimally processed signals \cite{gkikas_rojas_painformer_2025}. This capability has made deep learning increasingly attractive for physiological analysis. However, most existing deep learning approaches process physiological recordings as conventional time series, requiring the network to implicitly learn temporal dependencies from local observations. In short and noisy physiological recordings, these temporal relationships may be weakly expressed or obscured by inter-subject variability, limiting the model’s ability to capture coordinated dynamical behavior.

To address this limitation, we propose a framework that explicitly models the temporal dynamics of physiological signals using Hankel matrix embeddings. The Hankel transformation converts each time series into a structured trajectory representation composed of overlapping lagged observations, preserving temporal dependencies across multiple time scales \cite{isidori1985nonlinear,markovsky2012low}. Unlike conventional sequence-based representations, Hankel embeddings enable the signal to be analyzed as a coherent dynamical system, facilitating the learning of structured temporal patterns \cite{wang2017time}. This representation is particularly suitable for physiological signals such as BVP and RESP, where pain responses emerge from coordinated temporal behavior rather than isolated events. By integrating structured dynamical representations with deep learning, the proposed approach aims to model physiological pain responses without handcrafted features.  

Building on this representation, the proposed framework introduces three key contributions: 1) BVP and RESP signals are represented using Hankel matrix embeddings, enabling the explicit modelling of temporal dynamics and underlying physiological system behaviour beyond conventional sequence-based representations; 2) dedicated Fully Convolutional Networks (FCNs) are employed for each modality to automatically learn discriminative dynamical features directly from the structured Hankel representations without handcrafted feature engineering; and 3) a multimodal late fusion strategy is introduced to integrate complementary dynamical information from BVP and RESP for three-class classification of No Pain (NP), Low Pain (LP), and High Pain (HP). Overall, the proposed framework therefore provides a promising direction for scalable and objective pain assessment using multimodal physiological sensing.

\section{Proposed Framework} 
\label{sec:framework}
The goal of the proposed framework is to classify pain states by interpreting BVP and RESP signals as observable outputs of an underlying physiological process, rather than as independent stochastic sequences \cite{zanetti2019information}. The hypothesis is that pain induces a specific shift in the system's state-space trajectory. To capture this shift, rather than relying on handcrafted statistical features, our methodology uses a structural embedding: each signal is projected into a higher-dimensional trajectory representation via a Hankel matrix and processed by a Fully Convolutional Network (FCN). This design is motivated by realization theory \cite{isidori1985nonlinear} and by Takens' Embedding Theorem \cite{takens2006detecting}, which provide a basis for reconstructing informative state-space representations from time-delayed observations of a signal, in place of conventional flat-vector representations.

\subsection{The Hankel Operator}



For a discrete-time signal $s(k)$ containing $N$ samples, the standard Hankel
matrix $\mathbf{H}\in\mathbb{R}^{L\times K}$ is defined as

\begin{equation}
\label{eq:hankel}
\mathbf{H} =
\begin{bmatrix}
s(1)   & s(2)   & s(3)   & \cdots & s(K) \\
s(2)   & s(3)   & s(4)   & \cdots & s(K+1) \\
s(3)   & s(4)   & s(5)   & \cdots & s(K+2) \\
\vdots & \vdots & \vdots & \ddots & \vdots \\
s(L)   & s(L+1) & s(L+2) & \cdots & s(N)
\end{bmatrix},
\end{equation}

where $L$ is the embedding depth and $K=N-L+1$ is the number of shifted
state vectors. In this standard formulation, the matrix has anti-diagonal symmetry, such that $H_{i,j}$ is constant for a fixed value of $(i+j)$
\cite{isidori1985nonlinear}.

More generally, the delayed and strided trajectory representation used in the
hyperparameter search is defined element-wise as

\begin{equation}
\label{eq:generalised_hankel}
H_{i,j} =
s\left(1+(i-1)\tau+(j-1)r\right),
\end{equation}

where $\tau$ is the delay between successive elements of each state vector and
$r$ is the stride between successive state vectors. The number of valid
columns is therefore

\begin{equation}
K =
\left\lfloor
\frac{N-1-(L-1)\tau}{r}
\right\rfloor + 1.
\end{equation}

The standard Hankel matrix in \eqref{eq:hankel} is recovered when
$\tau=r=1$. More generally, anti-diagonal symmetry is retained when ($\tau=r$), whereas configurations with ($\tau\neq r$) produce a delayed and strided trajectory matrix that does not necessarily have constant anti-diagonals. In this work, the term generalized Hankel embedding refers to this broader trajectory-matrix formulation. The embedding depth ($L$) determines the temporal horizon, while ($\tau$) and ($r$) control the delay within each state vector and the stride between successive state vectors, respectively. These parameters were selected through the nested hyperparameter search described in Section~\ref{sec:evaluation_protocol}. Separate generalized Hankel matrices were constructed for BVP and RESP before modality-specific feature extraction and late fusion.


\subsection{FCN on Hankel-Embedded Inputs}



The Hankel matrix is provided to the FCN as an input with $L$ channels and $K$
temporal positions. Convolution is performed along the column, or time-shift,
dimension. For output feature map $q$, the activation at position $j$ is

\begin{equation}
G_{q,j} =
\sigma\left(
b_q+
\sum_{i=0}^{L-1}
\sum_{m=0}^{M-1}
W_{q,i,m}H_{i,j+m}
\right),
\end{equation}

where $M$ is the kernel length, $W_{q,i,m}$ is the convolutional weight
associated with input channel $i$, kernel position $m$, and output feature map
$q$, and $b_q$ is the corresponding bias. This operation jointly combines
information across delayed channels while learning local patterns along the
temporal evolution of the embedded trajectory.



Architectural implementation and training details are provided in Sec.~\ref{sec:methods}.



\section{Methodology}\label{sec:methods} 

\subsection{Dataset Organization and Preprocessing}
This study utilizes BVP and RESP data from the AI4PAIN dataset, which includes recordings from 65 healthy volunteers (42 males and 23 females). Detailed information regarding the data acquisition protocol, stimulation procedure, and experimental setup can be found in \cite{fernandez2024ai4pain,fernandez2025ai4pain}. BVP and RESP signals were recorded at a sampling frequency of 100 Hz. During the experiment, a 60-second baseline period was recorded to represent the NP condition. Pain stimulation was delivered using electrical stimuli at two intensity levels relative to each participant: LP and HP. For each intensity level, 12 stimulation trials of 10 seconds were recorded.


For this study, the recorded signals were segmented into 10-second epochs corresponding to each experimental condition. For every participant, six non-overlapping NP segments were extracted from the 60-second pre-stimulation baseline period, while 12 segments were obtained from each of the LP and HP stimulation conditions. To ensure a balanced class distribution, an additional six NP segments were obtained from the final 10 seconds of selected 40-second post-stimulation rest periods. The final 10-second interval was selected to avoid the immediate post-stimulation recovery period and reduce the potential influence of residual responses to the preceding pain stimulus. In total, the dataset comprised 2,340 observations, corresponding to 36 segments per participant (12 per class: NP, LP, and HP). The final dataset was therefore evenly balanced across the three experimental conditions, with 780 samples in each class. Prior to Hankel embedding, the BVP and RESP signals were preprocessed using a fourth-order Butterworth band-pass filter with frequency ranges of 0.5-5.0 Hz and 0.1-0.5 Hz, respectively, followed by z-score normalization. Z-score normalization was performed independently within each 10-second segment using only the samples contained in that segment. Consequently, the normalization of a test segment did not use information from any training
participant or from other test segments. Examples of the processed signals are shown in Figure~\ref{fig:bvp_resp}.

\begin{figure}[ht]
\centering
\includegraphics[width=1\linewidth]{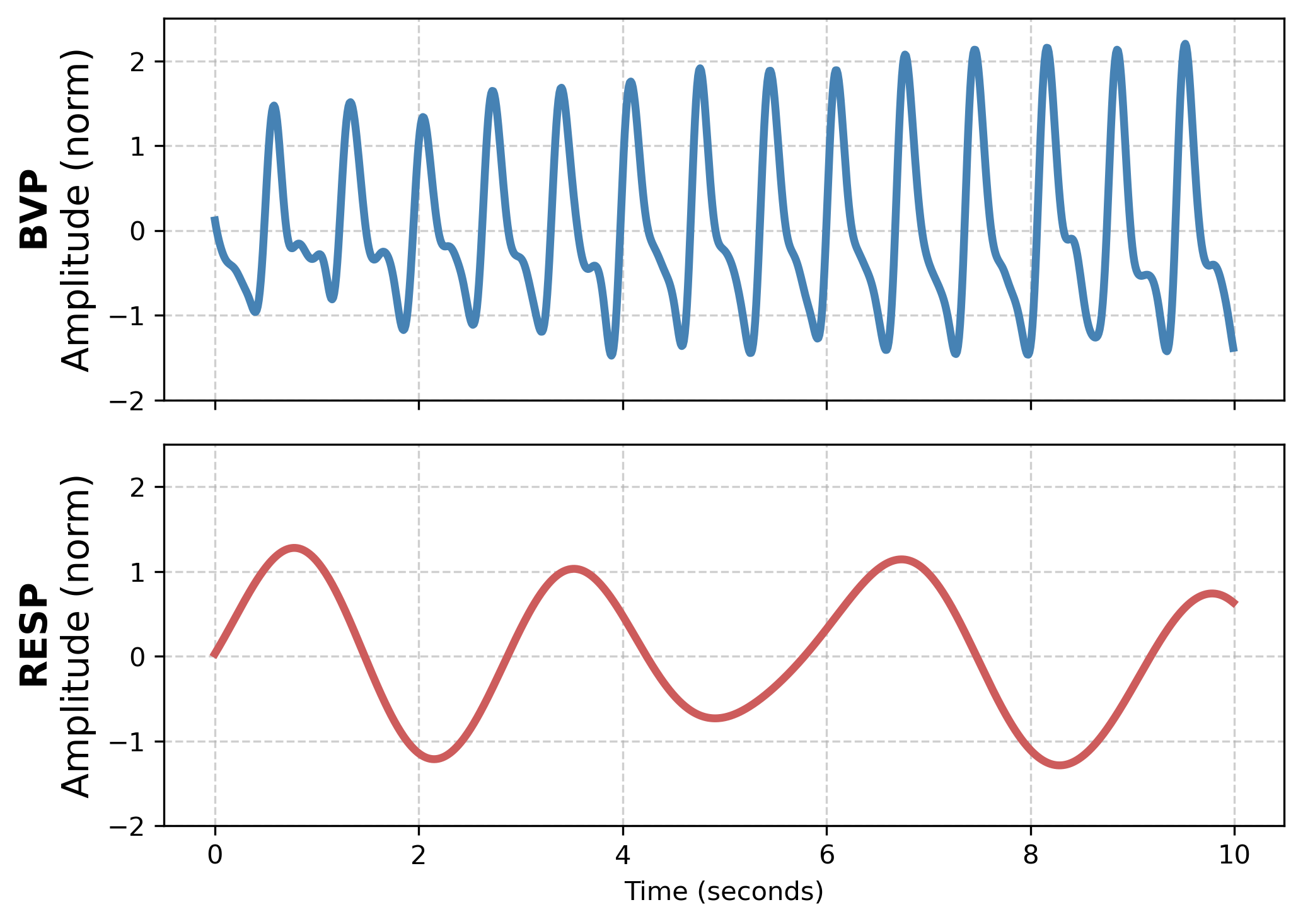}
\caption{Filtered and normalized BVP and RESP signals.}
\label{fig:bvp_resp}
\end{figure}

\subsection{Evaluation Protocol and Hyperparameter Optimization}
\label{sec:evaluation_protocol}

The proposed framework was evaluated using nested leave-one-subject-out cross-validation (LOSO-CV) to assess subject-independent generalization. In each outer fold, one participant was held out exclusively for testing, while the remaining participants were used for model training and hyperparameter optimization. Within the outer training set, 50 Optuna trials were conducted using subject-wise inner cross-validation, with participants kept separate between the training and validation sets. Hyperparameters were selected by maximizing the mean F1 score across the inner validation folds. The search included Hankel embedding parameters, convolutional depth, filter sizes, kernel sizes, activation functions, dropout rates, learning rate, weight decay, and classifier settings, as summarised in Table~\ref{tab:hpo_space}.

\begin{table}[b]
\centering
\caption{Hyperparameter Search Space for the Proposed Model.}
\label{tab:hpo_space}
\begin{tabular}{llll}
\hline
\textbf{Component} & \textbf{Hyperparameter} & \textbf{Search Space} \\
\hline

\multirow{2}{*}{General} 
& Learning Rate & $[10^{-4}, 5\times10^{-3}]$\\
& Weight Decay & $[10^{-8}, 10^{-3}]$ \\
\hline

\multirow{3}{*}{\makecell[l]{Hankel \\ Embedding}} 
& Depth & $\{10, 20, 30, 40, 50\}$ \\
& Delay & $\{1, 2\}$ \\
& Stride & $\{1, 2\}$ \\
\hline

\multirow{5}{*}{FCN} 
& Convolution Blocks & $\{1, 2, 3, 4\}$ \\
& Filters per Layer & $\{32, 64, 128, 256\}$ \\
& Kernel Size & $\{3, 5, 7, 9\}$ \\
& Activation Function & \{ReLU, ELU, GELU, Tanh\} \\
& Dropout & $[0.1, 0.5]$\\
\hline

\multirow{3}{*}{\makecell[l]{Classifier \\ Head}} 
& Hidden Units & $\{64, 128, 256\}$ \\
& Activation Function & \{ReLU, ELU, GELU, Tanh\} \\
& Dropout & $[0.1, 0.5]$ \\
\hline

\end{tabular}
\end{table}

Model training was performed using the Adam optimizer with cross-entropy loss and a batch size of 128. The learning rate and weight decay were selected independently by Optuna within each outer LOSO fold. Models were trained for a maximum of 500 epochs, with early stopping monitored on validation F1 score using a patience of 50 epochs. A random seed of 42 was used for reproducibility.

This procedure was repeated until each of the 65 participants had served once as the outer test subject, resulting in 65 independently optimized outer-fold models. Performance was reported as the mean and standard deviation across the 65 outer folds. Model performance was assessed using overall accuracy (Acc), macro-averaged F1-score, and class-wise true positive rate (TPR) for NP, LP, and HP. The class-specific TPR was calculated as the proportion of samples from each class that were correctly classified. All metrics were calculated separately for each held-out participant and summarised across the 65 outer LOSO folds. 

As hyperparameter optimization was performed independently within each outer fold, the resulting model configurations varied across folds. For architectural illustration, the 65 configurations were ranked according to their held-out accuracy, and the configuration corresponding to the median-performing fold was selected as a representative instance of the proposed Hankel-FCN. This selection was used solely for architectural illustration and did not influence hyperparameter optimization or the aggregate performance reported across the 65 outer folds.

\subsection{Hankel-FCN Framework}

The proposed framework constructs independent Hankel matrices for the BVP and RESP signals to represent each modality in a structured delay-coordinate space. These matrices preserve temporal dependencies and provide a dynamical representation of the physiological response to pain. For feature extraction, a dual-branch FCN architecture was implemented, with each branch processing one modality independently to enable modality-specific feature learning. Figure~\ref{fig:block_diagram} presents the representative configuration selected according to the procedure described in Section~\ref{sec:evaluation_protocol}.

Each convolutional block includes a Conv1D layer, Batch Normalization, a non-linear activation function, and Dropout regularisation. Global Average Pooling is applied after branch-specific feature extraction to obtain compact modality-specific feature vectors. These vectors are then combined through late fusion using concatenation, forming a joint representation that integrates complementary information from BVP and RESP.

For the representative configuration shown in Figure~\ref{fig:block_diagram}, the BVP branch contains three convolutional blocks, whereas the RESP branch contains one convolutional block. The fused representation is passed to a classification head containing a fully connected hidden layer with 256 units, GELU activation, Dropout of 0.4, and a final linear layer for three-class classification into NP, LP, and HP.

\begin{figure}[t]
\centering
\includegraphics[width=1\linewidth]{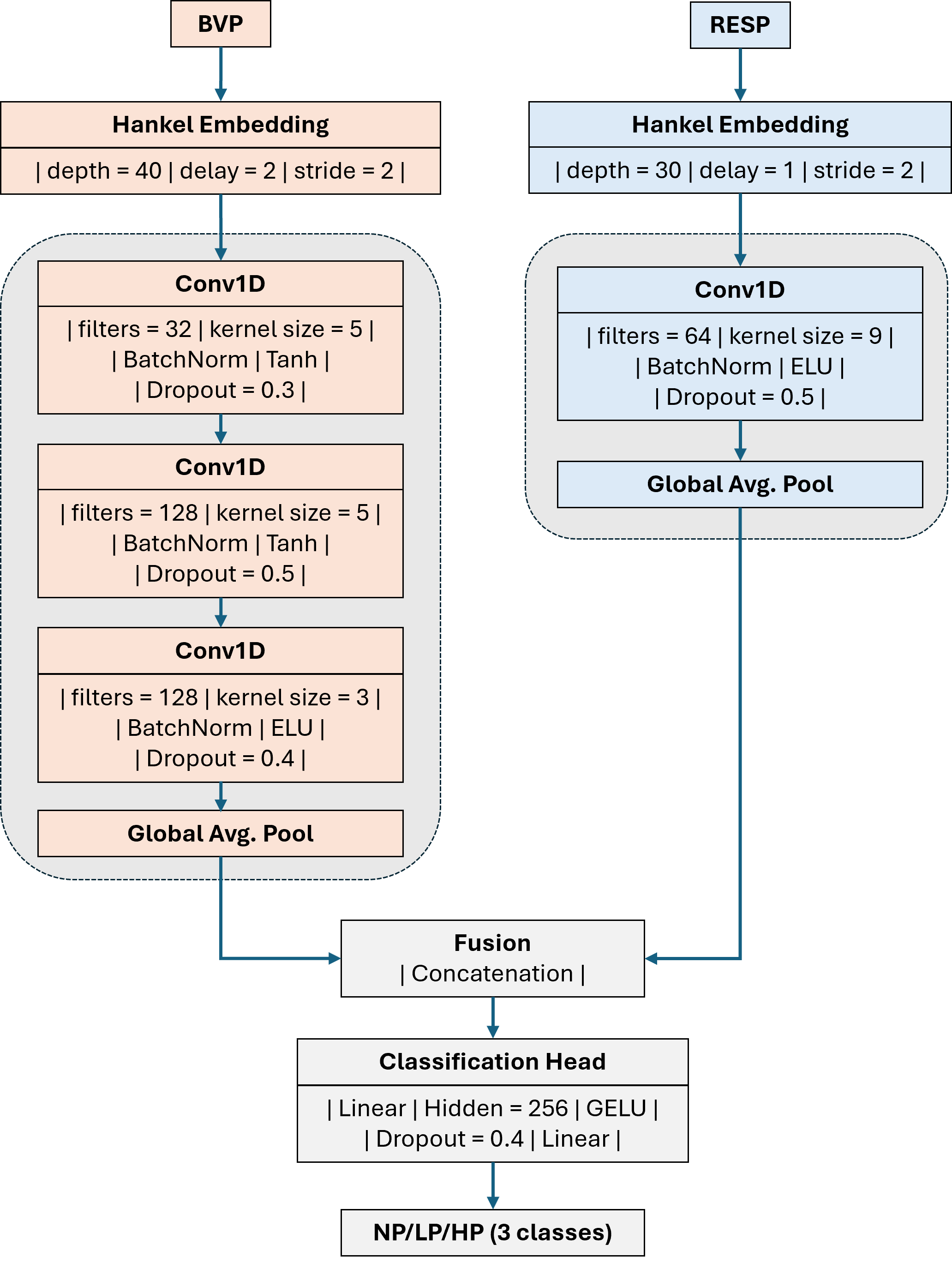}
\caption{Representative architecture of the proposed multimodal Hankel-FCN framework. Modality-specific Hankel embeddings and convolutional branches extract temporal features from BVP and RESP signals, which are fused for pain classification (NP, LP, HP).}
\label{fig:block_diagram}
\end{figure}


\section{Experimental Results} \label{sec:results}


We conducted multiclass classification experiments to distinguish NP, LP, and HP. The proposed Hankel-FCN was first evaluated using unimodal BVP, unimodal RESP, and multimodal BVP--RESP configurations. The multimodal framework integrated the modality-specific representations through the fusion block. We then compared multimodal FCN pipelines with and without Hankel embedding. All configurations were independently hyperparameter-optimized using the nested LOSO procedure described in Section~\ref{sec:evaluation_protocol}.  
The classification results are reported in Table~\ref{tab:results}, the Hankel comparison is presented in Figure~\ref{fig:hankel_ablation}, and the confusion matrix for the proposed multimodal Hankel-FCN is shown in Figure~\ref{fig:conf_matrix}.

As shown in Table~\ref{tab:results}, unimodal models yielded limited performance. While the BVP-based model achieved an accuracy of $58.08\% (\pm 13.90)$, it exhibited limited discriminative capability across pain intensities. Similarly, the RESP-based model achieved an overall accuracy of $52.74\% (\pm 8.87)$, indicating a relatively weak ability to capture pain-related patterns when considered in isolation. These findings highlight the inherent limitations of single-modality approaches, where physiological signals may be insufficiently informative or affected by noise and inter-subject variability to reliably characterize complex nociceptive states.


In contrast, the proposed multimodal approach, which integrates cardiovascular and respiratory dynamics (refer to Figure~\ref{fig:block_diagram}), demonstrates a substantial improvement in performance. The combined model achieves an accuracy of $67.09\%$ and an F1-score of $66.41\%$, indicating enhanced capability in distinguishing between pain levels. This improvement suggests that the integration of complementary physiological modalities enables a more comprehensive representation of the underlying autonomic response, leading to more robust and reliable pain classification.


\begin{table}[t]
\centering
\caption{Performance of the proposed Hankel-FCN model under unimodal and multimodal input configurations.}
\label{tab:results}
\resizebox{1\columnwidth}{!}{ 
\begin{tabular}{lccccc}
\toprule
\textbf{Modality} & \textbf{Acc (\%)} & \textbf{F1 (\%)} & \textbf{TPR (NP)} & \textbf{TPR (LP)} & \textbf{TPR (HP)} \\
\midrule
BVP 
& 58.08 $\pm$ 13.90 
& 56.89 $\pm$ 14.97 
& 62.95 $\pm$ 22.58 
& 56.28 $\pm$ 23.48 
& 55.00 $\pm$ 18.74 \\

RESP 
& 52.74 $\pm$ 8.87 
& 48.03 $\pm$ 11.24 
& 64.36 $\pm$ 31.23 
& 50.64 $\pm$ 28.58 
& 43.21 $\pm$ 29.82 \\

BVP, RESP 
& 67.09 $\pm$ 12.87 
& 66.41 $\pm$ 13.61 
& 68.68 $\pm$ 19.50 
& 64.74 $\pm$ 19.64 
& 68.08 $\pm$ 18.49 \\
\bottomrule
\end{tabular}
}
\end{table}

The Hankel-FCN achieved higher average values across the reported metrics. As shown in Figure~\ref{fig:hankel_ablation}, the independently optimized Hankel-FCN achieved a mean accuracy of 67.09\%, compared with 62.56\% for the non-Hankel FCN. The corresponding macro-F1 scores were 66.41\% and 61.85\%, respectively. For the Hankel and non-Hankel configurations, the class-specific TPRs were 68.68\% and 66.41\% for NP, 64.74\% and 56.92\% for LP, and 68.08\% and 64.36\% for HP, respectively. This comparison suggests that the structured temporal representation provided by the Hankel embedding may offer an additional benefit beyond multimodal fusion.



\begin{figure}[t]
\centering
\includegraphics[width=\linewidth]{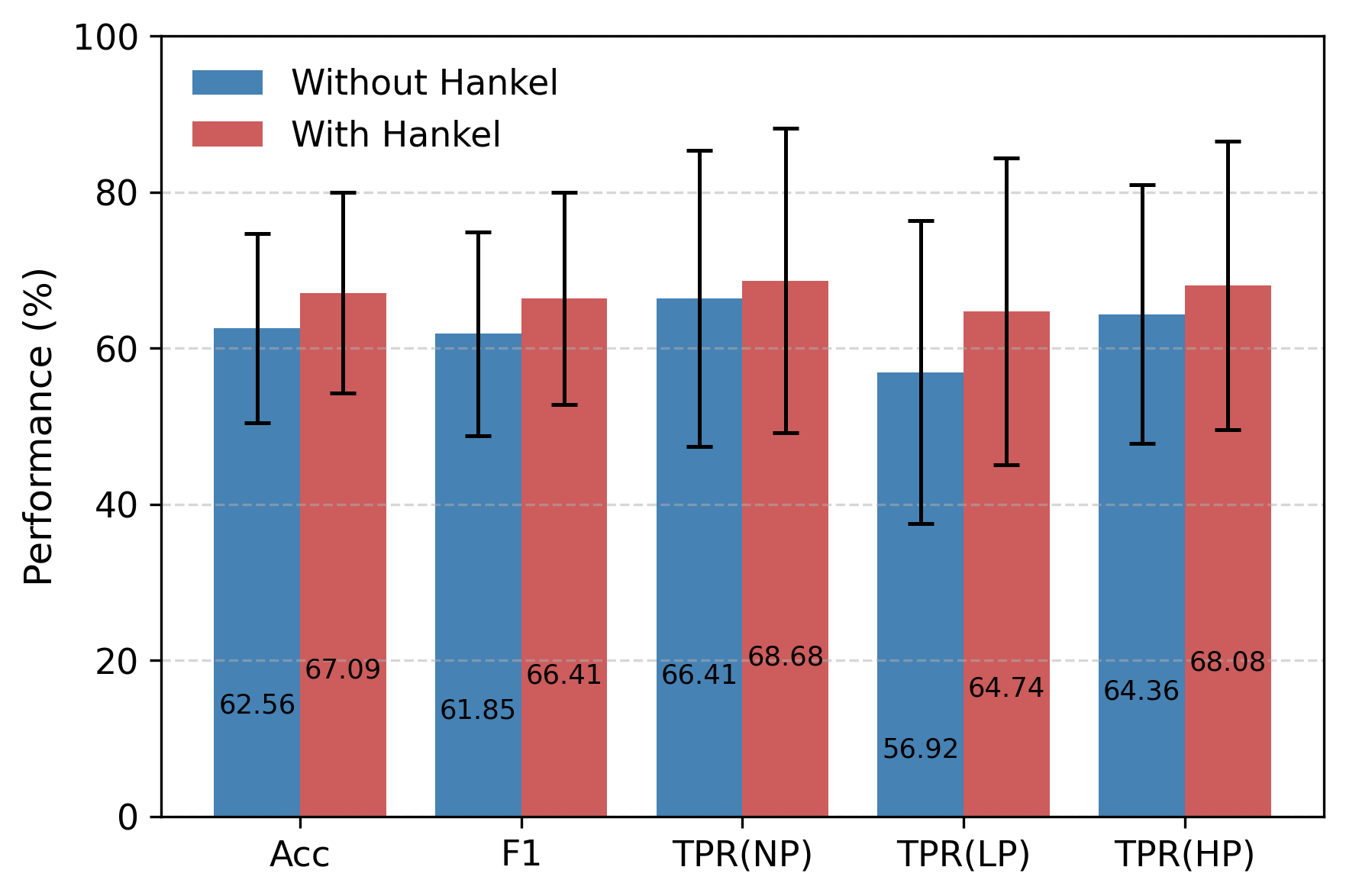}
\caption{Comparison of the independently optimized multimodal FCN pipelines with and without Hankel embedding across the 65 outer LOSO folds. Bars show the mean, and error bars indicate the standard deviation across held-out participants.}
\label{fig:hankel_ablation}
\end{figure}



The proposed model showed relatively balanced performance across the three pain
classes. The confusion matrix (Figure~\ref{fig:conf_matrix}) shows true positive
rates (TPRs) of 68.68\%, 64.74\%, and 68.08\% for NP, LP, and HP, respectively.
LP had the lowest TPR, with misclassifications distributed between the NP and
HP classes. The comparable TPRs for NP and HP suggest that the model identified
these classes with similar reliability, whereas LP was more difficult to
distinguish. This pattern may reflect overlapping physiological responses across
pain intensities, particularly for the intermediate LP condition.

Despite this inherent overlap, the multimodal framework demonstrates improved discrimination compared to unimodal approaches, reducing misclassifications across pain levels. This result highlights the benefit of integrating complementary physiological dynamics, which provide additional context for distinguishing subtle variations in autonomic responses. Overall, the findings suggest that combining cardiovascular and respiratory information enables a more robust representation of pain-related states, supporting more reliable classification in the presence of inter-class similarity.

In addition, to investigate Hankel hyperparameter selection patterns across subjects, we analyzed the configurations selected across the 65 outer LOSO folds. For BVP, the most frequently selected embedding depth, delay, and stride were $L=40$ (35/65 folds), $\tau=2$ (41/65), and $r=1$ (34/65), respectively. For RESP, the corresponding values were $L=10$ (20/65), $\tau=1$ (35/65), and $r=2$ (36/65). These findings reveal modality-dependent preferences in Hankel embedding configurations, while the variation in parameter selection across folds suggests sensitivity to inter-subject variability, particularly for RESP embedding depth.



\begin{figure}[t]
\centering
\includegraphics[width=0.8\linewidth]{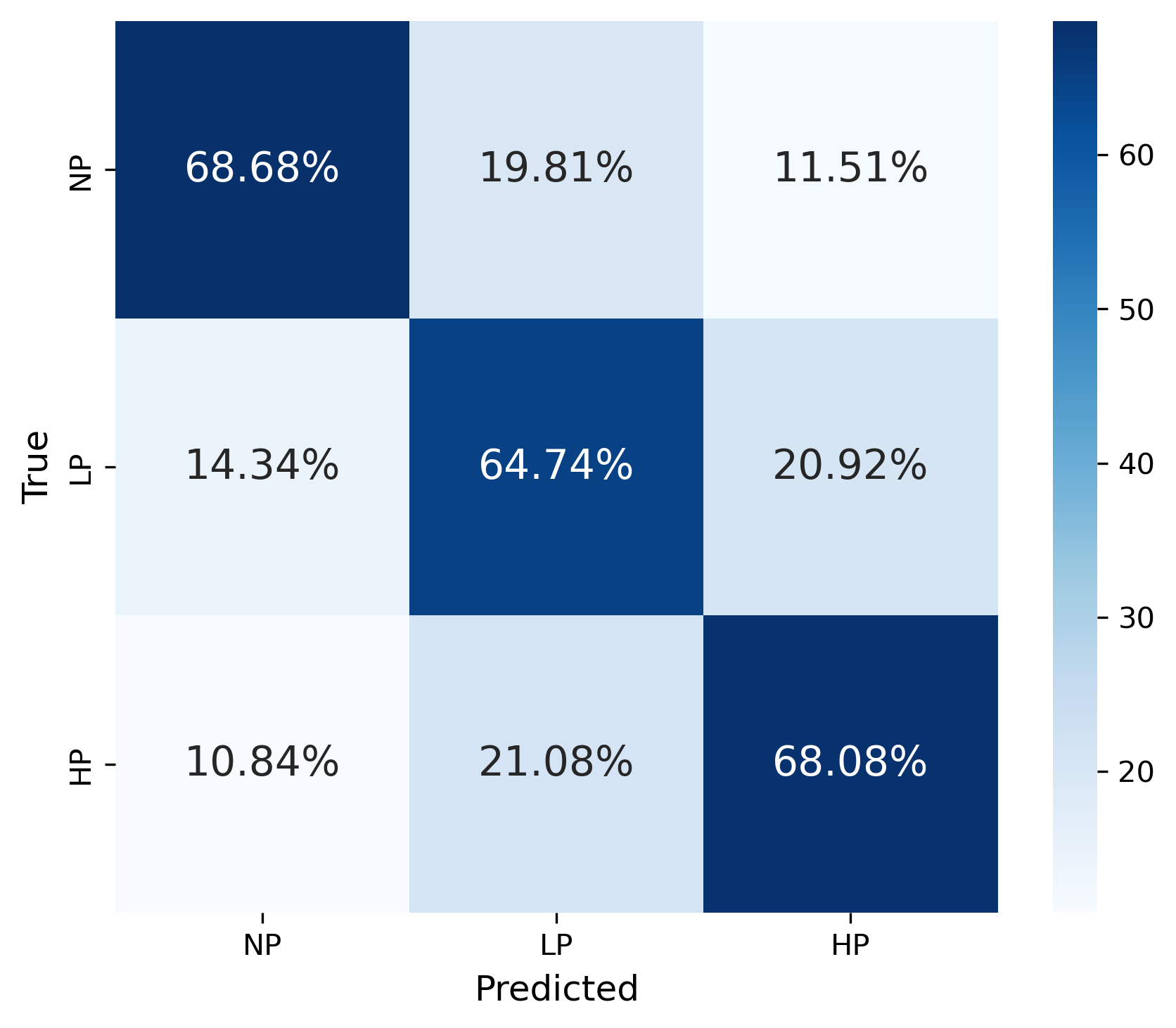}
\caption{Mean confusion matrix across the 65 outer LOSO folds for three-class pain classification (NP, LP, HP).}
\label{fig:conf_matrix}
\end{figure}

\subsection{Statistical Analysis}

Two complementary statistical analyses were conducted across the 65 LOSO-CV iterations. The first compared the multimodal BVP--RESP configuration with the unimodal BVP and RESP configurations to evaluate the benefit of multimodal fusion. The second analysis compared the independently optimized multimodal Hankel-FCN and non-Hankel FCN pipelines. Normality was evaluated for the participant-level paired accuracy differences using the Jarque--Bera and Anderson--Darling tests. Neither test identified significant departures from normality, and no extreme outliers were observed. Two-sided paired $t$-tests were therefore used, with Cohen's $d_z$ to quantify effect size and Holm's procedure to control the family-wise error rate across the three comparisons: multimodal BVP--RESP versus BVP, multimodal BVP--RESP versus RESP, and Hankel versus non-Hankel.

Multimodal fusion significantly improved accuracy compared with either
unimodal configuration. The multimodal BVP--RESP configuration achieved a mean accuracy of 67.09\% (SD = 12.87\%), compared with 58.08\% (SD = 13.90\%) for BVP and 52.74\% (SD = 8.87\%) for RESP. Two-sided paired $t$-tests showed that the multimodal configuration improved accuracy over BVP by 9.02 percentage points
(95\% CI: 5.19--12.84; $t(64)=4.71$; $d_z=0.58$; Holm-adjusted $p=2.78\times10^{-5}$) and over RESP by 14.36 percentage points
(95\% CI: 10.51--18.21; $t(64)=7.45$; $d_z=0.92$; Holm-adjusted $p=9.18\times10^{-10}$). These results indicate that combining BVP and RESP provided a consistent advantage over either modality alone.

The Hankel transformation produced a small improvement in accuracy. The Hankel configuration achieved a mean accuracy of 67.09\% (SD = 12.87\%), compared with
62.56\% (SD = 12.11\%) for the non-Hankel configuration. This represented a
mean improvement of 4.53 percentage points (95\% CI: $-0.07$--9.13; two-sided
paired $t$-test, $t(64)=1.97$; $d_z=0.24$; Holm-adjusted $p=0.053$). The exact paired sign-flip permutation test produced the same conclusion. Although the estimate favored the Hankel configuration, the evidence was not conclusive. Further validation is therefore required to determine whether the observed improvement is robust.

Overall, multimodal fusion was the primary source of performance improvement.
The contribution of the Hankel transformation was small and more variable
across participants, requiring further validation in a larger independent
cohort.

\section{Discussion}\label{sec:discussion}
The primary objective of this study was to evaluate the effectiveness of our proposed framework for objective pain assessment using BVP and RESP signals. Our results demonstrate that combining these physiological modalities provides more reliable pain classification than relying on individual signals. Although BVP and RESP signals operate on different physical scales, following a state-space augmentation via the Hankel matrix, their integration offers complementary information regarding autonomic nervous system responses to pain.

The findings clearly support multimodal fusion. Combining BVP and RESP produced significantly higher accuracy than either modality alone. The independently optimized Hankel-FCN also achieved higher average accuracy than the non-Hankel FCN, although the difference was smaller and more variable across participants. Because the two pipelines were independently hyperparameter-optimized, the observed difference may reflect both the Hankel representation and differences in the selected model configurations. A controlled ablation that holds the architecture and training hyperparameters constant would be required to isolate the specific contribution of the Hankel transformation. The current findings should therefore be interpreted as preliminary evidence that a Hankel-based pipeline may complement multimodal fusion.


Class-level performance was relatively balanced. As evidenced in Table~\ref{tab:results}, the NP and HP classes are identified with comparable TPRs of $68.68\%$ and $68.08\%$, respectively, while the LP class shows a slightly lower TPR of $64.74\%$. The lower TPR for LP indicates that the intermediate pain condition was more difficult to distinguish, potentially because of overlapping physiological responses between adjacent pain intensities. Nevertheless, the multimodal configuration achieved higher class-specific TPRs than either unimodal configuration across all three classes, supporting the complementary contribution of BVP and RESP.

The proposed Hankel-FCN achieved competitive performance against previous
AI4Pain studies. As shown in Table~\ref{tab:ai4pain_full_comparison}, the
proposed BVP--RESP model achieved an accuracy of 67.09\% using three-class classification and LOSO-CV. This result was slightly higher than the 66.55\% reported for the fNIRS-based SVM model \cite{khan2024multilevel}, and the 66.40\% reported for the EDA-based SVM model \cite{khan2026pain}. The proposed model also exceeded the 59.60\% achieved by the BVP-based ensemble \cite{khan2026pain} and improved on the 52.30\% reported in the multimodal baseline study \cite{fernandez2025ai4pain} by 14.79 percentage points. However, these comparisons are descriptive because the studies used different sensing modalities, preprocessing procedures, model configurations, and implementation settings. The results therefore indicate that the proposed approach performs competitively under the LOSO-CV protocol, without implying statistical superiority over previously published methods.

\begin{table}[t]
\caption{Comparison with previous studies using the AI4Pain dataset for three-class pain classification using LOSO-CV.}
\label{tab:ai4pain_full_comparison}
\centering
\scriptsize
\setlength{\tabcolsep}{3pt}
\resizebox{\columnwidth}{!}{%
\begin{tabular}{l l l c}
\toprule
\textbf{Ref.} & \textbf{Modality} & \textbf{Model} & \textbf{Acc. (\%)} \\
\midrule
\cite{fernandez2025ai4pain}\textsuperscript{1,b}    & EDA, BVP, RESP, SpO$_2$ & SVM (RBF)           & 52.30 \\
\cite{khan2026pain}\textsuperscript{b}     & EDA   & SVM (RBF)   & 66.40 \\
\cite{gkikas2024twins}\textsuperscript{a}    & fNIRS \& Video   & Twins-Pain ViT   & 46.46 \\
\cite{khan2026giaformer}\textsuperscript{a,b}       & EDA                              & GIAFormer     & 51.62 \\
\cite{khan2024multilevel}\textsuperscript{a}    & fNIRS   & SVM (RBF)   & 66.55 \\
\cite{khan2026pain}\textsuperscript{b}  & BVP     & Ensemble  & 59.60 \\
This work\textsuperscript{b}                      & BVP, RESP                        & Hankel-FCN    & 67.09 \\
\bottomrule
\end{tabular}%
}
\vspace{1mm}
{\footnotesize \textsuperscript{1}Baseline paper. AI4Pain release: \textsuperscript{a}2024 (fNIRS \& Facial Video),\\
\textsuperscript{b}2025 (Physiological).\par}
\end{table}

It is also worth situating the Hankel-based embedding relative to other temporal representations commonly used in physiological time-series modeling. Sliding-window approaches, which segment the signal into fixed-size local patches, are computationally simpler but leave the network to implicitly relearn continuity between adjacent windows, since the explicit ordering of temporal shifts is not preserved across patches. Direct time-delay embedding with a small, fixed set of lagged channels retains short-range dependencies but omits the exhaustive, overlapping structure of a full trajectory matrix. The representation adopted in this work can be regarded as a generalized Hankel or trajectory-matrix embedding. In the conventional case, overlapping lagged vectors form a matrix with constant anti-diagonals. In the more general delayed and strided formulation used in the hyperparameter search, this symmetry is retained only when the delay and stride are equal. When these parameters differ, the representation remains a structured trajectory matrix but is not a conventional Hankel matrix. In both cases, the formulation allows the convolutional branches to jointly process current and delayed observations across the selected temporal horizon, distinguishing it from the fixed, non-overlapping local patches used by conventional sliding-window approaches.


Finally, the use of FCN, combined with local normalization and a late fusion strategy, enables effective handling of modality-specific variability while preserving critical temporal information. This design allows each modality to be processed independently before integration, facilitating the learning of complementary physiological patterns without being dominated by a single signal source. As a result, the framework is designed to capture the underlying dynamics associated with different pain states. Overall, multimodal fusion was the main driver of performance, while structured dynamical representations remain a promising but not yet conclusive component.


Despite the promising results, several limitations should be acknowledged. First, the framework is evaluated on a single publicly available dataset, and further validation on larger and more diverse cohorts is needed to assess generalisability. Second, although only the final 10 seconds of the 40-second post-stimulation rest periods were used to obtain the additional NP observations, residual effects of the preceding stimulation cannot be completely excluded. Differences in recording context between the NP and stimulation conditions may therefore have contributed to the learned representations. Third, while the proposed approach leverages structured dynamical representations, the interpretability of the learned features remains limited. Finally, the current study focuses on classification performance and does not explore aspects such as computational efficiency, real-time deployment, or clinical integration, which are important considerations for practical applications.

\section{Conclusion and Future Work}\label{sec:conclusions}

This study presented a deep learning framework for three-class pain
classification using BVP and RESP signals. The results showed that multimodal
fusion provided a consistent improvement over either modality alone, achieving
an accuracy of 67.09\% and an F1 score of 66.41\% under nested LOSO-CV. The
Hankel representation produced a smaller descriptive improvement over the
non-Hankel FCN, but this effect was not conclusive and requires further
validation. Overall, the findings support multimodal physiological modeling
as the primary source of improvement and identify structured temporal
embedding as a promising direction for further investigation.

Future work will focus on improving discrimination between LP and HP states while evaluating the proposed framework on additional datasets, including those targeting stress and other internal human states \cite{gkikas2026unwind} that may influence pain perception or its manifestation. Furthermore, we aim to investigate richer physiological representations, including joint multivariate Hankel embeddings of BVP and RESP and additional modalities such as EDA and SpO$_2$~\cite{kyprakis_gkikas_acii_2026, gkikas2025tiny}. Complementary behavioral information, including facial expressions~\cite{gkikas_reface_acii_2026} and body movements~\cite{cruz_gkikas_emotion_acii_2026}, could also be explored to improve generalization and sensing reliability. Finally, future research will explore strategies to reduce computational complexity and support real-time pain assessment~\cite{gkikas_cruz_pain_eeite_2026}, facilitating potential applications in clinical monitoring systems and emotion-aware companion robots~\cite{montiel_cruz_gkikas_eeite_2026} within healthcare environments.

\bibliographystyle{ieeetr}
\bibliography{references.bib}

\end{document}